\documentclass[letterpaper]{article} % DO NOT CHANGE THIS
\usepackage{main}  % DO NOT CHANGE THIS
\usepackage{times}  % DO NOT CHANGE THIS
\usepackage{helvet}  % DO NOT CHANGE THIS
\usepackage{courier}  % DO NOT CHANGE THIS
\usepackage[hyphens]{url}  % DO NOT CHANGE THIS
\usepackage{graphicx} % DO NOT CHANGE THIS
\usepackage{natbib}  % DO NOT CHANGE THIS AND DO NOT ADD ANY OPTIONS TO IT
\usepackage{caption} % DO NOT CHANGE THIS AND DO NOT ADD ANY OPTIONS TO IT
\usepackage{algorithm}
\usepackage{algorithmic}
\usepackage{bm}
\usepackage{amsmath}
\usepackage{amssymb}
\usepackage{multirow}
\usepackage{etoolbox}

\usepackage{newfloat}
\usepackage{listings}
\DeclareCaptionStyle{ruled}{labelfont=normalfont,labelsep=colon,strut=off} % DO NOT CHANGE THIS
\floatstyle{ruled}
\newfloat{listing}{tb}{lst}{}
\floatname{listing}{Listing}
\nocopyright %-- Your paper will not be published if you use this command
\title{PersonaVlog: Personalized Multimodal Vlog Generation with Multi-Agent Collaboration and Iterative Self-Correction}
\author{
    Xiaolu Hou\textsuperscript{\rm 1}\equalcontrib, 
    Bing Ma\equalcontrib, 
    Jiaxiang Cheng, 
    Xuhua Ren \footnotemark[4], 
    Kai Yu, 
    Wenyue Li, \\
    Tianxiang Zheng, 
    Qinglin Lu\footnotemark[4]
}
\affiliations{
    \textsuperscript{\rm 1} Fudan University
    
}
\usepackage{bibentry}
\begin{document}

\maketitle
\renewcommand{\thefootnote}{\fnsymbol{footnote}}
\footnotetext[4]{Corresponding authors.}

\begin{abstract}
With the growing demand for short videos and personalized content, automated Video Log (Vlog) generation has become a key direction in multimodal content creation. Existing methods mostly rely on predefined scripts, lacking dynamism and personal expression. Therefore, there is an urgent need for an automated Vlog generation approach that enables effective multimodal collaboration and high personalization. To this end, we propose PersonaVlog, an automated multimodal stylized Vlog generation framework that can produce personalized Vlogs featuring videos, background music, and inner monologue speech based on a given theme and reference image. Specifically, we propose a multi-agent collaboration framework based on Multimodal Large Language Models (MLLMs). This framework efficiently generates high-quality prompts for multimodal content creation based on user input, thereby improving the efficiency and creativity of the process. In addition, we incorporate a feedback and rollback mechanism that leverages MLLMs to evaluate and provide feedback on generated results, thereby enabling iterative self-correction of multimodal content. We also propose ThemeVlogEval, a theme-based automated benchmarking framework that provides standardized metrics and datasets for fair evaluation. Comprehensive experiments demonstrate the significant advantages and potential of our framework over several baselines, highlighting its effectiveness and great potential for generating automated Vlogs.

\end{abstract}

\begin{links}
    \link{Project Page}{https://personavlog-paper.github.io/}
\end{links}

\section{Introduction}
With the rapid rise of short video platforms such as Douyin and Kuaishou, the era where “anyone can become a content creator” has arrived. In this new media landscape, video logs (Vlogs) have quickly become a mainstream and highly influential form of content. This is primarily due to their unique advantages in personal storytelling, creative expression, and showcasing individual style. Vlogs enable creators to vividly share their daily lives and unique perspectives. Through authentic and engaging storytelling, Vlogs can also establish deeper emotional connections with audiences. Therefore, developing a system that can automatically generate personalized and engaging Vlogs not only lowers the barriers to content creation, encourages broader user participation, but also holds significant practical value.

Previous methods \cite{kong2024hunyuanvideo,yang2024cogvideox,wan2025wan,gao2025seedance} primarily focus on generating short video clips of around five seconds. Although recent approaches such as Framepack \cite{zhang2025framepacking} and Self Forcing \cite{huang2025selfforcing} have started to explore long-form video generation, they still face significant challenges in maintaining narrative coherence, character consistency, and high-fidelity adherence to semantic prompts.
To address these issues, several frameworks \cite{xu2025mm, zhuang2024vlogger, xie2024dreamfactory, huang2025filmaster, huang2023freebloom, li2023videochat, wu2023visualchatgpt, zhang2023controllable, zhu2023minigpt} have introduced multi-agent systems, achieving notable improvements in narrative ability and shot composition. However, these frameworks have two key limitations. First, they rely heavily on detailed, high-quality pre-written inputs such as scripts or storyboards, which is inconsistent with the goal of automated and personalized Vlogs generation for large-scale users. Second, they typically adopt multi-stage agent pipelines. 
Inherent issues with generative models, such as semantic hallucinations and visual artifacts, often accumulate across multiple stages, ultimately resulting in low completion rates and suboptimal video quality.

To systematically address the above challenges, we propose PersonaVlog, an automated multimodal stylized Vlog generation framework that can produce personalized Vlogs featuring videos, background music, and inner monologue speech based on a given theme and reference image. Our main contributions include the following three aspects:
(\romannumeral1) \textbf{A Multimodal Multi-Agent Collaborative Framework (MACF) for Content Creation.} We propose MACF, which employs specialized agents to simulate a human production team. Based on Multimodal Large Language Models (MLLMs), these agents collaboratively accomplish coherent subtasks such as story generation, storyboarding, detailed video descriptions, character monologues, and music descriptions.
(\romannumeral2) \textbf{Multimodal Generation Feedback and Rollback Mechanism (FRM).} To enhance output quality, we introduce a closed-loop FRM, where MLLMs act as automated reviewers to provide feedback on generated keyframes and videos to enable self-correction. FRM effectively prevents quality deterioration and significantly improves the reliability and quality of the final results.
(\romannumeral3) \textbf{A Theme-based Vlog Automated Generation Evaluation Benchmark.} To facilitate reproducible research and fair comparison, we introduce ThemeVlogEval—the first comprehensive benchmark for this task. It features a wide range of topics, reference images, and style prompts, along with standardized, multidimensional evaluation metrics, thus providing a robust foundation for future advancements.
(\romannumeral4) Comprehensive experiments demonstrate the significant potential and advantages of our framework compared with several baselines, highlighting the effectiveness and great potential of our method for generating automated Vlogs. \textbf{We will release our code, benchmark, and models to support further research in the community.}

\section{Related Work}
\textbf{Long Video Generation.} Generating minute-long videos typically relies on parallel \cite{yin2023nuwa} or auto-regressive structures \cite{zhang2025framepacking,huang2025selfforcing,villegas2022phenaki}, but still faces significant challenges in narrative coherence, character consistency, and semantic alignment. Recently, LLM-based agents have attracted considerable attention. Studies show that multi-agent collaboration can improve the quality of long video generation \cite{hu2024storyagent,huang2025filmaster,zhuang2024vlogger,xie2024dreamfactory,wu2025automated}. However, existing approaches heavily rely on LLMs for text processing and are overly dependent on input scripts, resulting in limited creativity and diversity. To address these issues, we propose the multimodal Multi-Agent Collaboration Framework (MACF), which takes user-defined themes and personalized character images as input. By leveraging multi-agent collaboration and a review mechanism, MACF enables automatic and diverse content generation.

\noindent \textbf{Keyframe Image Generation.} Existing methods \cite{li2024photomaker, wang2024instantid, ye2023ip, tao2025instantcharacter} mainly enhance identity retention via parameter-efficient fine-tuning or large-scale pre-training, but are limited to human faces and training data, which restricts their generalization to non-human subjects. Training-free approaches \cite{zhou2024storydiffusion, tewel2024training} reduce training costs but often incur high memory and computation overhead, with limited consistency gains. Visual memory modules \cite{rahman2023make} improve contextual coherence in story generation, yet struggle with full-body consistency and lack effective feedback. Recent image edit models \cite{labs2025flux} advance subject consistency and flexible editing. Therefore, we leverage image edit model for keyframe image generation and introduce agent-based Feedback and Rollback Mechanism (FRM) to further enhance consistency and quality.

\section{Methodology}
\subsection{Overall Framework}
\begin{figure*}[t]
\centering
\includegraphics[width=0.98\textwidth]{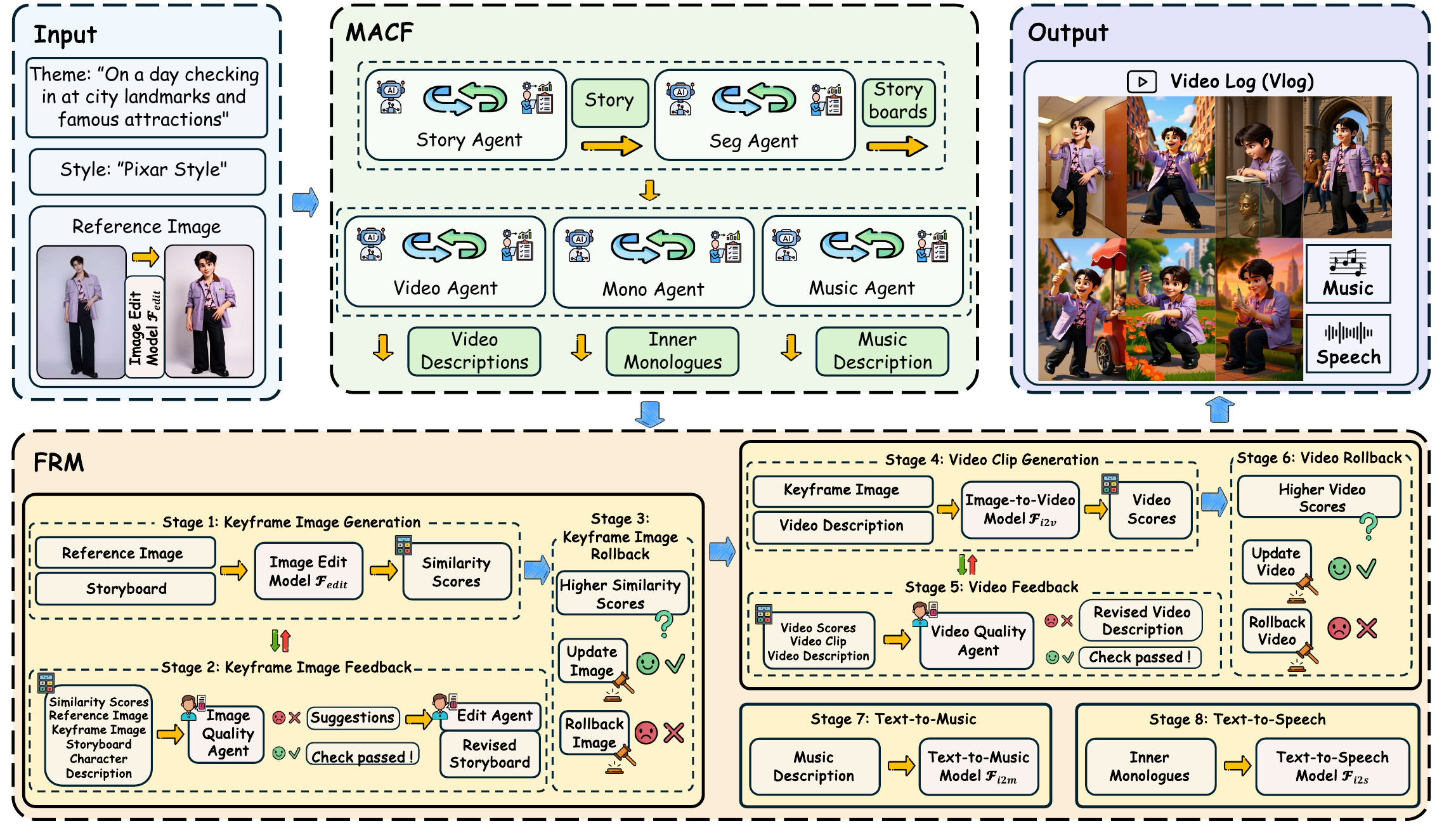}
\caption{\textbf{The overall framework of the proposed PersonaVlog}. PersonaVlog is based on a multimodal Multi-Agent Collaborative Framework (MACF) that can automatically generate complete and interesting stories, storyboards, video descriptions, character inner monologues, and background music descriptions based on input themes, character reference images, and styles. Subsequently, a Feedback and Rollback Mechanism (FRM) is used to iteratively optimize the multimodal generation results, ultimately achieving the automated generation of narrative, coherent, and diverse Video Log (Vlog) content.} 
\label{fig:pipeline}
\end{figure*}

As shown in Figure \ref{fig:pipeline}, given a theme $T_{theme}$, style $T_{style}$, and a real reference image $\bm{I}_r$, PersonaVlog aims to generate coherent and diverse narrative stylized Vlogs. Our PersonaVlog is a framework compatible with multiple backbone models. The workflow is as follows:
(\romannumeral1) The input real reference image is stylized by an image edit model $\mathcal{F}_{edit}$ to obtain a stylized reference image $\bm{I}_s$.
(\romannumeral2) Multi-Agent Collaborative Framework (MACF) utilizes multimodal multi-agent collaboration and review mechanisms to automatically generate multiple text descriptions based on the stylized character reference image and theme.
(\romannumeral3) Multiple backbone models are used to implement multimodal generation, including image-to-image, image-to-video, text-to-music, and text-to-speech.
(\romannumeral4) Feedback and Rollback Mechanism (FRM) is used to iteratively optimize multimodal generation results. 
Ultimately, PersonaVlog enables the automated generation of Vlogs with coherent narratives and diverse content.

\subsection{Multimodal Multi-Agent Collaborative Framework for Content Creation}
Based on a systematic analysis of theme-driven Vlog content, we carefully divide the constituent elements of a Vlog into five categories: complete and interesting stories, segmented storyboards, detailed video descriptions, inner monologues of character, and background music descriptions. To achieve coherent storytelling and diverse content in automatically generated Vlogs, it is necessary to conduct thorough divergent thinking and in-depth exploration of themes and their constituent elements.
Previous methods \cite{wu2025automated, xu2025mm} employ multi-agent collaboration paradigms to make progress in multi-scene video generation. However, they still have certain limitations: (\romannumeral1) they rely solely on Large Language Models (LLMs) for text processing, which is highly dependent on input scripts; (\romannumeral2) they lack the dynamic, personalized inner expressions commonly found in Vlogs. These issues result in generated content lacking creativity and diversity. 
To address the above problems, we propose the multimodal Multi-Agent Collaboration Framework (MACF) for content creation. MACF is user-centric, supporting input in the form of a theme and personalized character reference image. By combining multimodal information through multi-agent collaboration and review mechanisms, MACF automatically generates complete stories, storyboards, detailed video descriptions, character inner monologues, and music descriptions. 
Therefore, MACF can be customized according to individual characteristics and preferences, enabling the creation of more personalized, diverse, and appealing content.

Specifically, we construct a multi-agent collaboration and review mechanisms consisting of five types of agents: story agent $\mathcal{A}_1$, seg agent $\mathcal{A}_2$, video agent $\mathcal{A}_3$, mono agent $\mathcal{A}_4$, and music agent $\mathcal{A}_5$. Each type of agent $\left\{ \mathcal{A}_i \right\}_{i=1}^5$ implements two distinct roles: generator $\mathcal{G}_i$ and reviewer $\mathcal{R}_i$. Therefore, the system comprises ten agent instances with distinct functionalities $\left \{ \mathcal{G}_{i=1}^5, \mathcal{R}_{i=1}^5 \right \} $. Each agent instance is defined by a multimodal large language model (MLLM) and is associated with a prompt template. 

% See \textit{Supplementary Materials} for detailed descriptions and design specifications of all agents.

\noindent \textbf{Story Agent.}
Given a theme $T_{\text{theme}}$ and a stylized reference image $\bm{I}_s$, the task of the story agent is to perform comprehensive and divergent reasoning on both the theme and character elements. Generate a character description $T_\text{c}$ as well as a complete, engaging, and temporally coherent story $T_{\text{story}}$. 
Specifically, the generator $\mathcal{G}_1$ is responsible for generating $T_c$ and $T_{\text{story}}$, while the reviewer $\mathcal{R}_1$ evaluates whether $T_c$ and $T_{\text{story}}$ satisfy the requirements.
The above process can be modeled as follows:
\begin{equation}
(T_c, T_{\text{story}}) = \mathcal{R}_1 \circ \mathcal{G}_1 (T_{\text{theme}}, \bm{I}_s),
\end{equation}
where $\circ$ denotes the multi-round generate-review process.

\noindent \textbf{Seg Agent.}
Given a story $T_{\text{story}}$, the seg agent decomposes it into storyboards $\left\{ T_{\text{board}}^i \right\}_{i=1}^k$ with character behavior lines, where $k$ is the number of storyboards.
The generator $\mathcal{G}_2$ produces $\left\{ T_{\text{board}}^i \right\}_{i=1}^k$, and the reviewer $\mathcal{R}_2$ evaluates its compliance.
The above process can be modeled as follows:
\begin{equation}
 \left\{ T_{\text{board}}^i \right\}_{i=1}^k = \mathcal{R}_2 \circ \mathcal{G}_2 (T_{\text{story}}).
\end{equation}

\noindent \textbf{Video Agent.}
Given storyboards $\left\{ T_{\text{board}}^i \right\}_{i=1}^k$, the video agent's task is to generate more detailed video descriptions $\left\{ T_{\text{video}}^i \right\}_{i=1}^k$ based on these storyboards, including information such as camera movement, emotional atmosphere and cultural details.
Among them, the generator $\mathcal{G}_3$ produces $\left\{ T_{\text{video}}^i \right\}_{i=1}^k$, while the reviewer $\mathcal{R}_3$ is responsible for evaluating whether the video descriptions satisfy the requirements.
The above process can be modeled as follows:
\begin{equation}
 \left\{ T_{\text{video}}^i \right\}_{i=1}^k = \mathcal{R}_3 \circ \mathcal{G}_3 (\left\{ T_{\text{board}}^i \right\}_{i=1}^k).
\end{equation}

\noindent \textbf{Mono Agent.}
Given storyboards $\left\{ T_{\text{board}}^i \right\}_{i=1}^k$, the task of the mono agent is to generate character inner monologues $\left\{ T_{\text{mono}}^i \right\}_{i=1}^k$, reflecting their thoughts and feelings at the moment.
Among them, the generator $\mathcal{G}_4$ is responsible for generating $\left\{ T_{\text{mono}}^i \right\}_{i=1}^k$, while the evaluator $\mathcal{R}_4$ is responsible for evaluating whether the monologues satisfies the requirements.
The above process can be modeled as follows:
\begin{equation}
 \left\{ T_{\text{mono}}^i \right\}_{i=1}^k = \mathcal{R}_4 \circ \mathcal{G}_4 (\left\{ T_{\text{board}}^i \right\}_{i=1}^k).
\end{equation}

\noindent \textbf{Music Agent.}
Given storyboards $\left\{ T_{\text{board}}^i \right\}_{i=1}^k$ and a theme $T_{\text{theme}}$, the music agent generates a music description $T_{\text{music}}$ specifying the desired emotions, instruments, and styles for the background music.
Among them, the generator $\mathcal{G}_5$ is responsible for generating $T_{\text{music}}$, while the evaluator $\mathcal{R}_5$ is responsible for evaluating whether $T_{\text{music}}$ satisfies the requirements.
The above process can be modeled as follows:
\begin{equation}
 T_{\text{music}} = \mathcal{R}_5 \circ \mathcal{G}_5 (\left\{ T_{\text{board}}^i \right\}_{i=1}^k, T_{\text{theme}}).
\end{equation}

\subsection{Feedback and Rollback Mechanism for Multimodal Generation}
Previous methods \cite{wu2025automated} employ a two-stage generation paradigm. First, The text-to-image generation method \cite{, zhou2024storydiffusion} is used to generate images from storyboards, followed by image-to-video generation stage. 
Although these methods make progress in obtaining multi-scene videos, the following limitations remain: (\romannumeral1) poor subject consistency in keyframe images; (\romannumeral2) the absence of an effective multimodal feedback mechanism, resulting in limited visual diversity and consistency in the generated content. The above issues limit the quality and practicality of multimodal content generation.
To this end, we propose a multimodal content generation method based on the Feedback and Rollback Mechanism (FRM). This method comprises three main modules: image generation, video generation, and audio generation. Agent-based feedback is introduced in the image and video generation stages, providing feedback based on the generated content and its corresponding prompt, and combining multiple metrics to achieve automatic re-generation and rollback optimization. 
Audio content is automatically generated using text-to-music and text-to-speech models.
FRM can effectively identify and correct unreasonable issues in keyframe images and videos, ensuring that the final results are not inferior to the original state through the rollback strategy, thereby improving the overall quality and consistency of the keyframe images and videos.
This multimodal content generation method can reduce reliance on human intervention and manual screening to some extent, thereby effectively improving the diversity and consistency of multimodal content.

\noindent \textbf{Keyframe Image Generation.}
As shown in Figure \ref{fig:pipeline}, given stylized reference image $\bm{I}_s$ and storyboards $\left\{ T_{\text{board}}^i \right\}_{i=1}^k$ , we use the image edit model $\mathcal{F}_{edit}$ to obtain the keyframe images $\{ \bm{I}_\text{key}^i \}_{i=1}^k$.
This process can be defined as follows:
\begin{equation}
 \{ \bm{I}_\text{key}^i \}_{i=1}^k = \mathcal{F}_{edit}(\left\{ T_{\text{board}}^i \right\}_{i=1}^k, \bm{I}_s).
\end{equation}

\noindent \textbf{Keyframe Image Feedback and Rollback.}
In the keyframe image feedback mechanism, we construct a system consisting of two agents, the image quality agent $\mathcal{A}_\text{quality}^\text{image}$ and the edit agent $\mathcal{A}_\text{edit}$, to collaboratively perform quality assessment and optimization of keyframe images. Each agent instance is defined by a multimodal large language model and is associated with a prompt template.
For each keyframe image $\bm{I}_\text{key}^i$, we calculate its image-to-image feature similarity $m_\text{i2i}$ with the reference image $\bm{I}_s$, and its image-to-text feature similarity $m_\text{i2t}$ with the corresponding storyboard description $T_{\text{board}}^i$. These similarity scores serve as the basis for subsequent evaluation and rollback.

During the quality assessment stage, the image quality agent $\mathcal{A}_\text{quality}^\text{image}$ combines storyboard description $T_{\text{board}}^i$, character description $T_\text{c}$, character reference image $\bm{I}_s$, keyframe image $\bm{I}_\text{key}^i$ and similarity scores to determine whether the image has any of the following issues: (\romannumeral1) Incorrect number of limbs or body parts; (\romannumeral2) Abnormal expressions or poses; (\romannumeral3) Abnormal background or foreground; (\romannumeral4) Unreasonable clothing or accessories; (\romannumeral5) Insufficient resolution or clarity; (\romannumeral6) Inconsistent alignment of descriptions. If any of the above problems exist, the image quality agent directly outputs the modification suggestion $\hat{T}_\text{mod}^\text{i}$. Otherwise, the image is considered to be of qualified quality.
The above process can be formally modeled as:
\begin{equation}
\hat{T}_\text{mod}^i =
\begin{cases}
\mathcal{A}_\text{quality}^\text{image}(\bm{x}), & \text{if a problem exists},  \\
\varnothing, & \text{otherwise},
\end{cases}
\end{equation}
where $\bm{x} = (T_{\text{board}}^i,\, T_\text{c},\, \bm{I}_s,\, \bm{I}_\text{key}^i,\, m_\text{i2i},\, m_\text{i2t})$ denotes the set of input variables for the image quality agent.
For unqualified images, the edit agent $\mathcal{A}_\text{edit}$ converts $\hat{T}_\text{mod}^i$ into an accurate prompt suitable for image editing. This prompt is passed as input to image edit model $\mathcal{F}_\text{edit}$ together with the reference image $\bm{I}_s$ and storyboard $T_{\text{board}}^i$ to generate the optimized image $\bm{\hat{I}}_\text{key}^i$.
The above process can be formally modeled as:
\begin{equation}
    \bm{\hat{I}}_\text{key}^i = \mathcal{F}_\text{edit}(\mathcal{A}_\text{edit}(\hat{T}_\text{mod}^i), T_{\text{board}}^i, \bm{I}_s).
\end{equation}

Finally, recalculate the similarity scores $\hat{m}_\text{i2i}$ and $\hat{m}_\text{i2t}$ for $\bm{\hat{I}}_\text{key}^i$. If both scores are higher than the original scores $m_\text{i2i}$ and $m_\text{i2t}$, replace the original image with $\bm{\hat{I}}_\text{key}^i$. Otherwise, execute the rollback strategy and retain the original image.

\noindent \textbf{Video Generation}
As shown in Figure \ref{fig:pipeline}, given the keyframe image $\bm{\hat{I}}_\text{key}^i$ and video description $T_{\text{video}}^i$, we use the image-to-video generation model $\mathcal{F}_{i2v}$ to obtain the video clip $\bm{V}^i$. This process can be defined as follows:
\begin{equation}
    \bm{V}^i = \mathcal{F}_{i2v}(T_{\text{video}}^i, \bm{\hat{I}}_\text{key}^i).
\end{equation}

\begin{table*}[]
\centering
\renewcommand{\arraystretch}{1.05}
\resizebox{0.95\textwidth}{!}{%
\begin{tabular}{ccccccccccccc}
\hline
\multirow{2}{*}{ {Method}} & \multicolumn{4}{c}{Storyboard Metric   $\uparrow$} & \multicolumn{2}{c}{Image Metric   $\uparrow$} & \multicolumn{6}{c}{Video Metric   $\uparrow$} \\ \cline{2-13} 
 & SI & TC & BD & ThC & TIA & CC & SC & BC & MS & DyD & AQ & IQ \\ \hline
StoryDiffusion \cite{zhou2024storydiffusion} & - & - & - & - & 0.66 & 0.47 & - & - & - & - & - & - \\
InstantCharacter \cite{tao2025instantcharacter} & - & - & - & - & 0.79 & 0.49 & - & - & - & - & - & - \\
MM-StoryAgent \cite{hu2024storyagent} & 4.05 & 4.88 & 4.59 & 4.89 & 0.74 & 0.43 & - & - & - & - & - & - \\
MovieAgent \cite{xu2025mm} & 4.37 & 4.95 & 4.87 & 4.94 & 0.74 & 0.47 & 69.51 & 80.40 & 96.29 & 95.96 & 64.98 & 73.38 \\
\textbf{Ours (PersonaVlog)} & \textbf{4.57} & \textbf{4.99} & \textbf{4.98} & \textbf{4.97} & \textbf{0.79} & \textbf{0.53} & \textbf{82.95} & \textbf{87.81} & \textbf{97.52} & \textbf{99.69} & \textbf{67.49} & \textbf{73.85} \\ \hline
\end{tabular}%
}
\caption{\textbf{Performance comparison among the proposed framework and baselines.} ``SI'', ``TC'', ``BD'' and ``ThC'' refer to ``Story interest'', ``Temporal Continuity'', ``Behavioral Diversity'' and ``Thematic Consistency'' for storyboard metrics. ``TIA'' and ``CC'' refer to ``Text-Image Alignment'' and ``Character Consistency'' for keyframe image metrics. ``SC'', ``BC'', ``MS'', ``DyD'', ``AQ'' and ``IQ'' refer to ``Subject Consistency'', ``Background COnsistency'', ``Motion Smoothness'', ``Dynamic Degree'', ``Aesthetic Quality'' and ``Imaging Quality'' for video metrics. Our approach achieves the best results.}
\label{tab:cmp}
\end{table*}

\begin{table*}[]
\centering
\renewcommand{\arraystretch}{1.05}
\resizebox{0.95\textwidth}{!}{%
\begin{tabular}{ccccccccccccc}
\hline
\multirow{2}{*}{ {Method}} & \multicolumn{4}{c}{Storyboard Metric   $\uparrow$} & \multicolumn{2}{c}{Image Metric   $\uparrow$} & \multicolumn{6}{c}{Video Metric   $\uparrow$} \\ \cline{2-13} 
 & SI & TC & BD & ThC & TIA & CC & SC & BC & MS & DyD & AQ & IQ \\ \hline
GPT-4.1 (w/o multi agent \& FRM) & 4.53 & 4.87 & 4.85 & 4.88 & 0.75 & 0.51 & 80.88 & 87.50 & 97.24 & 96.79 & 66.36 & 73.36 \\
GPT-4.1 (w/ multi agent) & 4.57 & 4.99 & 4.98 & 4.97 & 0.78 & 0.52 & 81.05 & 87.56 & 97.32 & 99.00 & 67.33 & 73.78 \\
GPT-4.1 (w/ multi agent \& FRM-I) & 4.57 & 4.99 & 4.98 & 4.97 & 0.79 & 0.53 & 81.36 & 87.57 & 97.32 & 99.13 & 67.35 & 73.79 \\
GPT-4.1 (w/ multi agent \& FRM-V) & 4.57 & 4.99 & 4.98 & 4.97 & 0.78 & 0.52 & 81.96 & 87.63 & 91.41 & 99.47 & 67.38 & 73.82 \\
\textbf{Ours (PersonaVlog)} & \textbf{4.57} & \textbf{4.99} & \textbf{4.98} & \textbf{4.97} & \textbf{0.79} & \textbf{0.53} & \textbf{82.95} & \textbf{87.81} & \textbf{97.52} & \textbf{99.69} & \textbf{67.49} & \textbf{73.85} \\ \hline
\end{tabular}%
}
\caption{\textbf{Ablation results of different components.} FRM-I refers to the image feedback and rollback mechanism. FRM-V refers to the video feedback and rollback mechanism.}
\label{tab:ablation}
\vspace{-8pt}
\end{table*}

\noindent \textbf{Video Feedback and Rollback.}
In the video feedback mechanism, we define a video quality agent $\mathcal{A}_\text{quality}^\text{video}$ to perform video quality assessment and optimization. The agent instance is defined by a multimodal large language model and is associated with a prompt template.
For each video, we calculate video scores $m_\text{video}^i$, which assess imaging quality, subject consistency, background consistency, dynamic degree, motion smoothness, and aesthetic quality \cite{huang2024vbench}. These scores constitute the basis for subsequent evaluation and potential rollback.

During the quality assessment stage, the video quality agent $\mathcal{A}_\text{quality}^\text{video}$ combines video scores $m_\text{video}^i$, video description $T_{\text{video}}^i$ and video clip $\bm{V}^i$ to determine whether the video exhibits: (\romannumeral1) misalignment with the description; or (\romannumeral2) any anomalies (e.g., facial distortion, unreasonable motion, etc.). If any of these issues are detected, $\mathcal{A}_\text{quality}^\text{video}$ outputs a revised video prompt $\hat{T}_\text{video}^\text{i}$ and reason for modification $T_\text{reason}^i$. Otherwise, the video is considered qualified.
The above process can be formally modeled as:
\begin{equation}
\left \{ \hat{T}_\text{video}^\text{i}, T_\text{reason}^i \right \}=
\begin{cases}
\mathcal{A}_\text{quality}^\text{video}(\bm{y}), & \text{if a problem exists},  \\
\varnothing, & \text{otherwise},
\end{cases}
\end{equation}
where $\bm{y} = (T_{\text{video}}^i, \bm{V}^i, m_\text{video}^i)$ denotes the set of input variables for the video quality agent. 
For unqualified videos, the image-to-video model $\mathcal{F}_{i2v}$ receives the revised video prompt $\hat{T}_\text{video}^\text{i}$ and $\bm{\hat{I}}_\text{key}^i$ to regenerate the video clip $\bm{\hat{V}}^i$.
This process can be defined as follows:
\begin{equation}
    \bm{\hat{V}}^i = \mathcal{F}_{i2v}(\hat{T}_\text{video}^\text{i}, \bm{\hat{I}}_\text{key}^i).
\end{equation}

Finally, recalculate the video scores $\hat{m}_\text{video}$ for the new video clip $\bm{\hat{V}}^i$. If all scores are higher than the original scores, replace the original video with $\bm{\hat{V}}^i$. Otherwise, execute the rollback strategy and retain the original video.

\noindent \textbf{Audio Generation.}
In the automated generation of diverse Vlog content, background music and character audio are essential. We employ a text-to-music model $\mathcal{F}_{i2m}$ to generate background music $\mathcal{M}_{bgm}$ from the music prompt $T_{\text{music}}$, and a text-to-speech model $\mathcal{F}_{i2s}$ to synthesize the character’s inner monologue voice $\mathcal{M}_{sp}^{i}$ using $T_\text{mono}^i$ and reference audio $s$. This process is defined as follows:
\begin{equation}
    \mathcal{M}_{bgm} = \mathcal{F}_{i2m}(T_{\text{music}}), \mathcal{M}_{sp}^{i} = \mathcal{F}_{i2s} (T_\text{mono}^i, s).
\end{equation}

\begin{figure*}[t]
\centering
\includegraphics[width=0.82\textwidth]{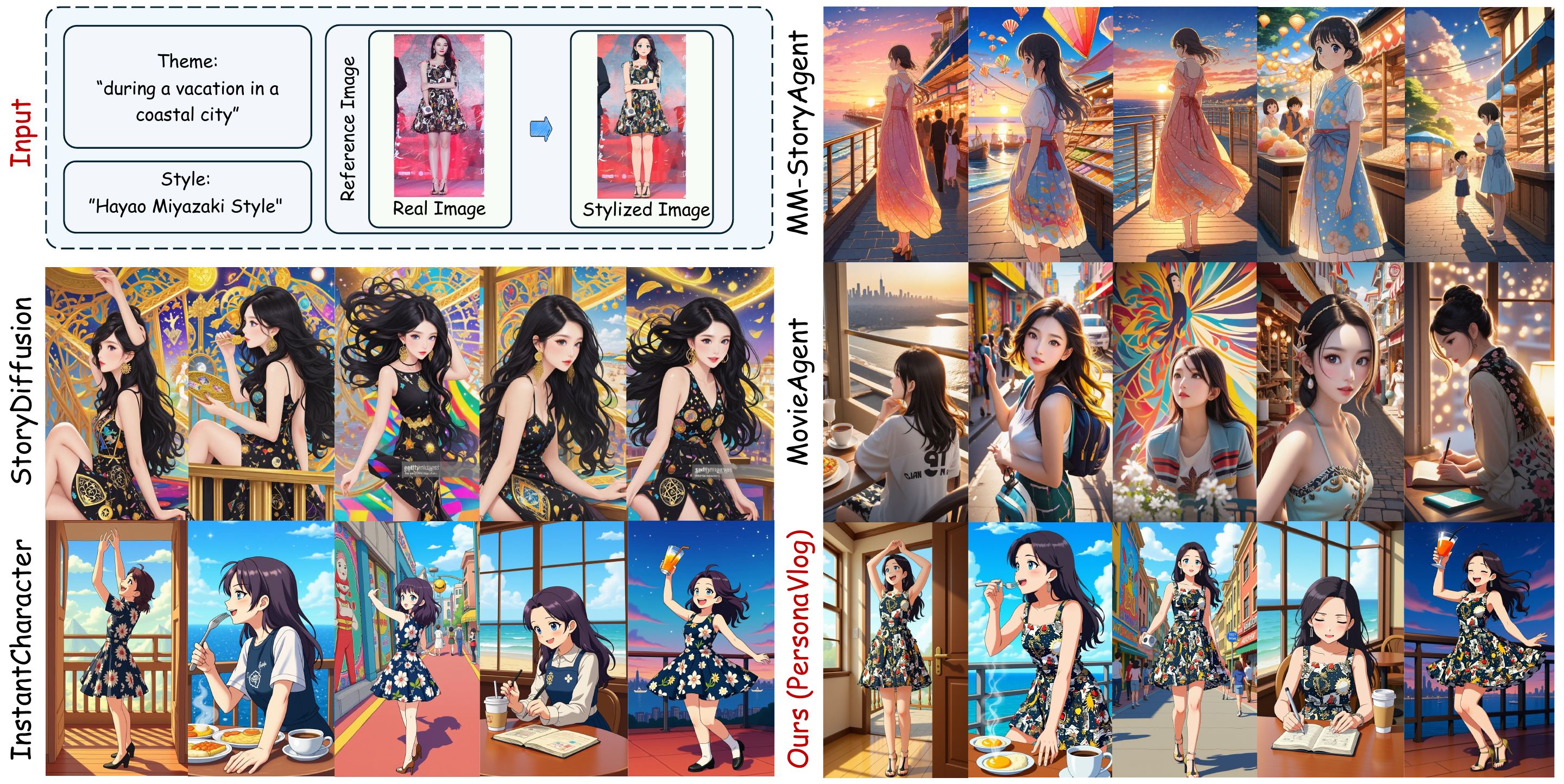}
\caption{\textbf{Qualitative Comparison of Our Method and Baselines.} On the Theme ``During a vacation in a coastal city'' in ``Hayao Miyazaki Style''.}
\label{fig1}
\vspace{-8pt}
\end{figure*}

\subsection{Theme-based Vlog Automated Generation Evaluation Benchmark}
Previous methods \cite{wu2025automated, zhou2024storydiffusion} mostly employ complete story scripts or randomly generated, irrelevant keyframe prompts from LLMs to evaluate the quality of the generated content. Although the above methods can measure the quality of the generated content to a certain extent, they still have the following limitations:
(\romannumeral1) Lack of image and video evaluation datasets driven by themes, styles, and reference images; (\romannumeral2) Lack of automated evaluation metrics applicable to multiple generation tasks. These issues limit the further advancement of theme- and reference image-driven stylized Vlogs in terms of unified comprehensive evaluation and continuous development.
To this end, we propose ThemeVlogEval, a theme-based automated benchmarking framework for evaluating Vlog content generation. Our approach addresses the limitations of previous methods, which lack evaluation datasets and metrics based on themes, styles, and reference images.

\noindent \textbf{Source Data.}
To ensure the diversity and representativeness of the benchmark dataset, we carefully design the selection of reference images, visual styles, and themes. For the reference images, we select 10 high-quality celebrity portraits with a balanced gender ratio (half male and half female), with each gender group containing four adults and one child. This selection strategy aims to cover the different age and gender characteristics commonly found in Vlogs, thereby enhancing the broad applicability of the evaluation. 
For visual style, we select two styles that are widely used and representative in Vlog content creation to evaluate the generated model under different style conditions.

Theme selection combines LLM-assisted generation with manual screening. We first use advanced LLM to generate a series of diverse daily Vlog themes, which are then manually screened and adjusted to ensure their relevance, diversity, and applicability to Vlog content generation tasks. This process ensures that the selected themes are representative of both realistic and surreal Vlog scenarios, while also posing a certain challenge to existing generative models.

\noindent \textbf{Evaluation Metrics.}
We combine MLLM and traditional visual multimodal models to simultaneously evaluate metric scores for storyboard, keyframe images, and videos.
In terms of storyboard evaluation, we use MLLM to score storyboards based on four dimensions: (\romannumeral1) \textbf{Story Interest}, i.e., whether the story content is interesting and appealing;  (\romannumeral2) \textbf{Temporal Continuity}, i.e., whether the story timeline is coherent and clear;  (\romannumeral3) \textbf{Behavioral Diversity}, i.e., whether the behaviors of the characters are rich and diverse; (\romannumeral4) \textbf{Thematic Consistency} i.e., the degree to which the story content aligns with the predefined theme. The model assigns scores of 1 to 5 for each dimension and provides the basis for the scores to ensure the explicability and objectivity of the assessment.
The process can be modeled as follows:
\begin{equation}
    \mathcal{M}(\left\{ T_{\text{board}}^i \right\}_{i=1}^k) = \left\{s_i, r_i\right\}_{i=1}^4,
\end{equation}
where $\mathcal{M}$ represents the MLLM, $s_i$ and $r_i$ is the score and reason for each dimension.

For keyframe image evaluation, we leverage state-of-the-art visual multimodal models to quantitatively assess generation results from two aspects: (\romannumeral1) \textbf{Text-Image Alignment}: The average CLIP score between each generated image and its corresponding storyboard script is used to measure semantic consistency. (\romannumeral2) \textbf{Character Consistency} $\mathcal{S}_\text{subj}$: This is calculated as a weighted average of the CLIP feature cosine similarity $\mathcal{S}_\text{clip}^\text{image}$ between generated and reference images, and the Euclidean distance $\mathcal{S}_\text{pose}$ of character skeleton keypoints among generated images. This metric comprehensively evaluates both identity preservation and the avoidance of excessive copy-paste, where higher scores indicate better subject consistency and greater pose diversity.
The process can be defined as follows:
\begin{equation}
    \mathcal{S}_\text{subj} = \alpha \times \mathcal{S}_\text{clip}^\text{image} + (1-\alpha) \times \mathcal{S}_\text{pose},
\end{equation}
where $\alpha$ denotes the weighting parameter, set to 0.5.

For video evaluation, we use the common video evaluation metrics \cite{huang2024vbench}. These include (\romannumeral1) \textbf{Subject Consistency}: evaluating whether the subject remains consistent throughout the video by assessing the similarity of dino features across frames; (\romannumeral2) \textbf{Background Consistency}: evaluating the temporal consistency of the background scene by assessing the similarity of CLIP features across frames; (\romannumeral3) \textbf{Motion Smoothness}: evaluating the smoothness of generated motion by leveraging motion priors from the video frame interpolation model; (\romannumeral4) \textbf{Dynamic Degree}: Using RAFT \cite{teed2020raft} to estimate the degree of motion in the synthesized video; (\romannumeral5) \textbf{Aesthetic Quality}: Evaluating the artistic and aesthetic value of each video frame; (\romannumeral6) \textbf{Imaging Quality}: Evaluating the distortion (e.g., overexposure, noise, blur) present in the generated frames.

\begin{figure*}[t]
\centering
\includegraphics[width=0.82\textwidth]{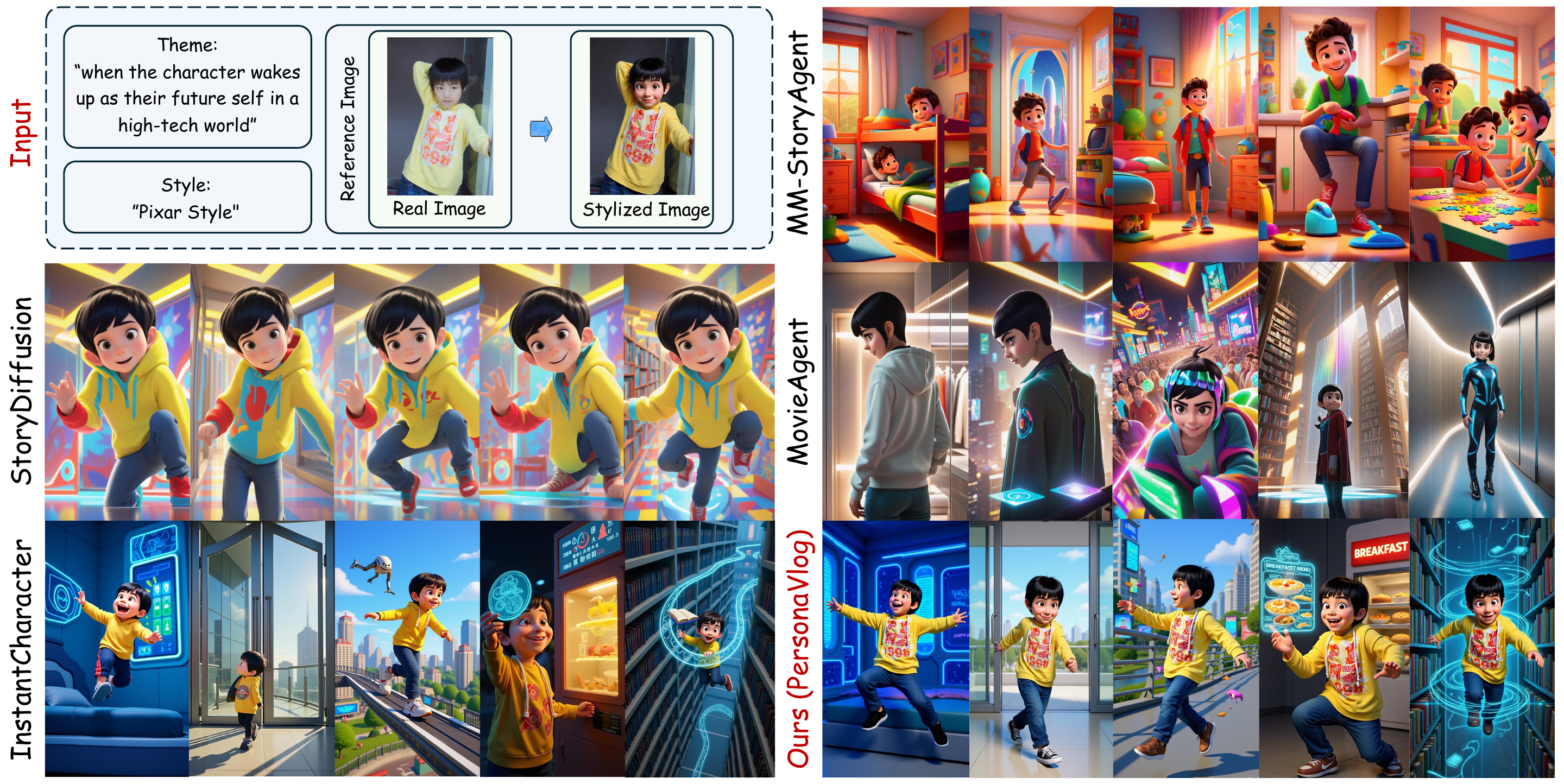}
\caption{\textbf{Qualitative Comparison of Our Method and Baselines.} On the Theme ``When the character wakes up as their future self in a high-tech world'' in ``Pixar Style''.}
\label{fig2}
\vspace{-6pt}
\end{figure*}

\section{Experiments}
\subsection{Experiment Setting}

Our proposed PersonaVlog can seamlessly integrate various MLLM architectures and base models. In the experiments, we use Google-gemini-2.5-pro \cite{comanici2025gemini} to construct the video quality agent and employ GPT-4.1 to construct the remaining agent instances. We use Flux-Kontext \cite{labs2025flux} as the image edit model, AudioX \cite{tian2025audiox} for text-to-music generation, CosyVoice \cite{du2024cosyvoice} for text-to-speech generation, and employ Wanx2.1 \cite{wan2025wan} for image-to-video generation. The framework's high compatibility and flexibility provide powerful support for the generation of multimodal content.
The inference steps for FLUX-Kontext are set to 20 steps, the classifier-free guidance scale is set to 3.5, and the image resolution is set to 768 × 1360. We carefully design the task templates and select high-quality context examples. All experiments are conducted on NVIDIA H20 GPUs.

\subsection{Comparison with State-of-the-Art Methods.}
We compare the proposed PersonaVlog with four representative and reproducible methods, including (\romannumeral1) subject consistency image generation methods: StoryDiffusion \cite{zhou2024storydiffusion} and InstantCharacter \cite{tao2025instantcharacter}; (\romannumeral2) script-driven video generation methods: MovieAgent \cite{wu2025automated}; (\romannumeral3) theme-driven image generation methods: MM-StoryAgent \cite{xu2025mm}. We use the open-source codebase of the above models to conduct a comprehensive evaluation on ThemeVlogEval. All models use stylized reference images processed by Flux-Kontext \cite{labs2025flux} as input, and the MACF module in PersonaVlog is used to convert the themes from ThemeVlogEval into the input formats required by each method.

\begin{figure}[t]
% \vspace{-10pt}
\centering
\includegraphics[width=1.0\columnwidth]{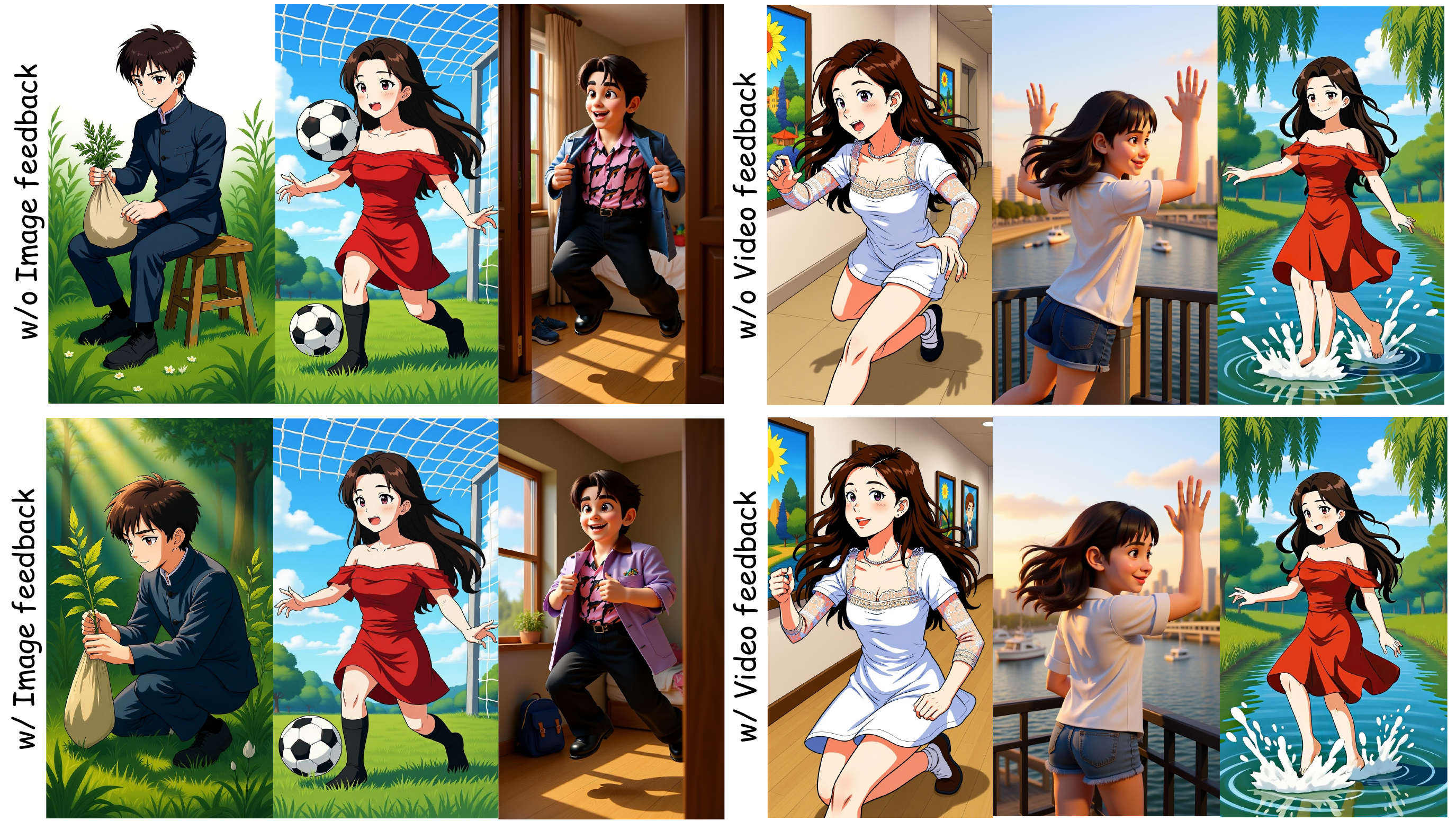}
% \vspace{-4pt}
\caption{Visualization of ablation results.}
\vspace{-10pt}
\label{fig3}
\end{figure}

\noindent \textbf{Qualitative Results.}
We conduct a qualitative analysis. Figure \ref{fig1} and \ref{fig2} show the visualization results generated by our method and baseline methods. We observe the following points:
(\romannumeral1) StoryDiffusion and MM-StoryAgent maintain style consistency across images, but show large style gaps from the reference and suffer from repetitive backgrounds and character poses.
(\romannumeral2) InstantCharacter provides diverse backgrounds and actions, but sometimes loses style or character identity (e.g., clothing, hairstyle, eye color).
(\romannumeral3) MovieAgent generates diverse images but struggles to maintain character consistency.
(\romannumeral4) In contrast, our method achieves better character consistency, style consistency, pose diversity, and narrative diversity.

\noindent \textbf{Quantitative Results.}
Table \ref{tab:cmp} shows the average quantitative results on ThemeVlogEval. We can conclude the following points:
(\romannumeral1) Overall, our method generates storyboards that surpass the baseline models in terms of coherence and diversity, while also yielding improved results in image and video quality.
(\romannumeral2) Our generated storyboards improved upon the best-performing baseline by 4.6\% and 2.3\% in terms of story interest and behavioral diversity respectively.
(\romannumeral3) Our keyframe images scored 12.8\% higher than the best-performing baseline in terms of character consistency, achieving better subject consistency and more diverse poses.
(\romannumeral4) Our method also maintains its lead in multiple video-related metrics.
These results demonstrate that the key components of our method can effectively facilitate in-depth exploration of themes and their elements, thereby enhancing the diversity and quality of images and videos.

\subsection{Ablation Studies.}
To validate the necessity of different components, we gradually add four modules: multi agent, image feedback and rollback, and video feedback and rollback. The results are shown in Figure \ref{fig3} and Table \ref{tab:ablation}. 
(\romannumeral1) Multi-agent collaboration significantly improves storyboard scores and has a positive impact on image and video scores. This indicates that multi-agent collaboration and review mechanisms help to explore themes and related elements in greater depth, thereby effectively improving content quality.
(\romannumeral2) The addition of image and video feedback and rollback mechanisms greatly improve generation quality by systematically evaluating key factors like image and background rationality, appearance consistency, and motion rationality.
(\romannumeral3) Integrating all modules can achieve optimal performance for all metrics, fully demonstrating the positive contributions of each module.

\section{Conclusion}
In this paper, we present PersonaVlog, a novel framework for automatic personalized Vlog generation. Our approach introduces a multimodal multi-agent collaborative framework, which orchestrates a set of specialized agents to systematically manage the entire Vlog creation pipeline, including story creation, storyboard segmentation, visual content generation, and background music composition. To ensure the generation of high-quality outputs, we embed a feedback and rollback mechanism for iterative self-correction within the framework. Furthermore, we propose ThemeVlogEval, a theme-based automated benchmarking framework, providing standardized metrics and datasets for a fair evaluation. Extensive experiments demonstrate the effectiveness of our framework. The results highlight the potential of PersonaVlog to facilitate more creative and high-quality multimodal content creation, paving the way for future research and applications in personalized media generation.

\bibliography{main}


\end{document}